\documentclass[10pt, a4paper]{article}
\usepackage{lrec-coling2024} 

\usepackage{helvet}

\usepackage{natbib}
\usepackage{multibib}
\makeatletter
\def\@mb@citenamelist{cite,citep,citet,citealp,citealt,citepalias,citetalias}
\makeatother
\newcites{languageresource}{~}

\usepackage{graphicx}
\usepackage{tabularx}
\usepackage{soul}
\usepackage{comment}
\usepackage{titlesec}

\usepackage{xcolor}
\usepackage{hyperref}
 \definecolor{darkblue}{rgb}{0, 0, 0.5}
  \hypersetup{breaklinks=true, colorlinks=true, citecolor=darkblue, linkcolor=darkblue, urlcolor=darkblue}

\usepackage[textwidth=20mm]{todonotes}
\setuptodonotes{size=\tiny}

\usepackage{tipa}

\usepackage{booktabs}
\usepackage[detect-all=true,
            round-mode=places,
            round-precision=1,
            round-pad=false,
            group-minimum-digits=4,
]{siunitx}

\title{Multilingual Power and Ideology Identification in the Parliament:\newline
  a Reference Dataset and Simple Baselines}

\name{Çağrı Çöltekin$^1$, Matyáš Kopp$^2$, Katja Meden$^{3,5}$,\\
      {\bf \large{}Vaidas Morkevicius$^4$, Nikola Ljubešić$^3$, Tomaž Erjavec$^3$}} 

\address{$^1$University of Tübingen, Tübingen, Germany,
         $^2$Charles University, Prague, Czech Republic,\\
         $^3$Jožef Stefan Institute, Ljubljana, Slovenia,
         $^4$Kaunas University of Technology, Kaunas, Lithuania,\\
         $^5$Jožef Stefan International Postgraduate School, Slovenia,\\
         ccoltekin@sfs.uni-tuebingen.de,
         kopp@ufal.mff.cuni.cz,
         katja.meden@ijs.si,\\
         vaidas.morkevicius@ktu.lt,
         nikola.ljubesic@ijs.si,
         tomaz.erjavec@ijs.si\\}

\abstract{
  We introduce a dataset on political orientation and power position identification.
  The dataset is derived from ParlaMint,
  a set of comparable corpora of transcribed parliamentary speeches from
  29 national and regional parliaments.
  We introduce the dataset,
  provide the reasoning behind some of the choices during its creation,
  present statistics on the dataset,
  and, using a simple classifier, some baseline results on predicting 
  political orientation on the left-to-right axis, 
  and on power position identification,
  i.e., distinguishing between the speeches delivered by governing coalition party members 
  from those of opposition party members.
 \\ \newline \Keywords{ideology, power, parliamentary corpus, ParlaMint} }

\begin{document}
\nocite{11356/1432}
\maketitleabstract

\section{Introduction}
Parliaments are one of the most important institutions in modern
democratic states where issues with high societal impact are discussed.
The decisions made in a national parliament
affect the citizens of its country
on fundamental aspects of their life.
The societal importance of parliamentary discourse requires
a better understanding and analysis of parliamentary debates.
As a result, there has been a recent increase in the number of resources
\citep{fiser2018-keyparla,CRF-parliamentary}
and (computational) linguistic analyses of parliamentary debates
\cite[see][for recent reviews]{glavas2019,abercrombie2020}.
The impact of the decisions made in a parliament often goes beyond their borders,
and may even have global effects.
Hence, comparative studies of parliamentary debates
across countries and in multiple languages
is also important. 

The dataset described here is derived from 
the ParlaMint corpora, 
a collection of comparable corpora of transcribed parliamentary speeches from
\num{29} national and regional parliaments,
covering at least the period from 2015 to 2022 \citep{erjavec2022}.
The dataset is prepared for a shared task on 
two important aspects of a political discourse,
\emph{political orientation} and \emph{power} \citep{kiesel2024}.%
\footnote{Further practical information about the shared task
can be found on the shared task web page at
\url{https://touche.webis.de/clef24/touche24-web/ideology-and-power-identification-in-parliamentary-debates.html}.}
Although a simplification,
political orientation on the left-to-right spectrum has been one of the defining
properties of political ideology \citep{arian1983,vegetti2019}.
Power is another factor that shapes the political discourse
\citep{vandijk2008,fairclough2013cda,fairclough2013lp}.
Despite its central role in critical discourse analysis,
to the best of our knowledge,
power was not studied computationally earlier.%
\footnote{Our definition of power for the present data set is also simplified.
  As suggested by an anonymous reviewer,
  other power roles,
  such as being a (shadow) cabinet member,
  or the role in the party may manifest differently in the speech.
  We leave such aspect of power in speech for future research.
}
We provide a reference dataset of parliamentary speeches for both tasks,
which we expect to be instrumental
for quantitative and computational studies
on ideology and power in parliamentary debates
beyond the present shared task as well.

Both tasks are formulated as binary classification tasks.
For the power position identification task, this choice is mostly straightforward,
as the distinction we want to make is
between the speeches delivered by governing party members
and those given by opposition party members.

Classifying political orientation is more complex, as it can be expressed in many ways.
In fact, ParlaMint provides annotations from two sources 
\citep{parlamint-metadata}:
Wikipedia and the Chapel Hill Expert Survey Europe \citep[CHES,][]{CHES}.
Wikipedia classifies the political orientations of parties
into 13 categories on the left-to-right spectrum,
as well as five other values that do not fit into this axis
(e.g., `Big Tent', or `Single Issue Politics' values).
Conversely, CHES gives political orientation along a large number of dimensions 
(\num{85} in total, e.g., stance towards European integration, but also the general left-to-right position of a party), 
with the numeric values based on averaged scores of expert surveys.
For the left-to-right position experts assigned a numeric score
between 0 to 10 (far left to far right)
based on a party's general ideological stance.
Not all parties have political orientation annotations in ParlaMint,
but the coverage of the Wikipedia annotations is more comprehensive than that of the CHES annotations.
As a result, we use orientation values from Wikipedia.

To facilitate graded predictions on the left-to-right scale,
we use labels 0 for left, and 1 for right-wing parties.
We mark Wikipedia categories from `far-left' (FL) to `centre to centre-left' (CCL) as \emph{left},
and those from `far-right' (FR) to `centre to centre-right' (CCR) as \emph{right}.
We exclude the speeches from the members of the parties marked as centre
and parties whose orientation does not fit into the left-to-right continuum.


For both tasks,
the main challenge in the creation of a dataset 
is to minimize the effects of covariates.
Even though the instances to classify are speeches,
the annotations are based on the party membership of the speaker.
As a result,
underlying variables like party membership,
or speaker identity perfectly covary
with ideology and power in most cases.
The sampling procedure described
in Section~\ref{sec:sampling} below aims to reduce these correlations,
and encourage systems trained on the data
to generalize to the particular task,
rather than predictions based on easier-to-guess covariates.

ParlaMint is a multilingual dataset of transcribed speeches delivered
in different regional and national parliaments.
As a result,
it also offers opportunities to investigate
similarities and differences of ideology and power
in varying cultures and parliamentary traditions,
as well as their reflection in different languages.
Even though the shared task does not offer a cross-lingual evaluation track,
the uniformly encoded data allows participants
to exploit `universal' aspects of ideology and power
through, for example, transfer learning.
To encourage participation in multiple languages,
and help participants build (simple) multilingual classifiers easily,
we also include automatic English translations of the speeches.

Our aim in this paper is to describe the process and rationale 
behind the dataset construction, as well as providing an overview
of the resulting data.
We also describe a trivial baseline
and the results of experiments with this baseline.

\section{Data}

\begin{table*}
  \centering
  \resizebox{\textwidth}{!}{\begin{tabular}{l*{4}{@{\hspace{5pt}}S[table-format=5.0,round-precision=0]%
                        @{\hspace{10pt}}S[table-format=2.1,round-precision=1]%
                        @{\hspace{10pt}}S[table-format=3.1,round-precision=1]}}
    \toprule
        & \multicolumn{6}{c}{Orientation} & \multicolumn{6}{c}{Power} \\ 
        \cmidrule(lr){2-7}\cmidrule(lr){8-13}
        & \multicolumn{3}{c}{Training} & \multicolumn{3}{c}{Test}
        & \multicolumn{3}{c}{Training} & \multicolumn{3}{c}{Test} \\ 
        \cmidrule(lr){2-4}\cmidrule(lr){5-7}\cmidrule(lr){8-10}\cmidrule(lr){11-13}
        & {n}& {L\%}& {tokens}
        & {n}& {L\%}& {tokens}
        & {n}& {O\%}& {tokens}
        & {n}& {O\%}& {tokens}\\
    \midrule
    Austria (AT)&               7879& 32.55&  535.4& 2002& 44.66&  566.6& 15971& 58.83&  568.1& 2181& 48.97&  598.5\\
    Bosnia and Herzegovina (BA)&1301& 20.91&  375.4& 2014& 28.90&  348.2&  2531& 16.83&  351.5& 1992& 16.92&  355.0\\
    Belgium (BE)&               2276& 32.07&  403.9& 2018& 38.21&  378.4&  4765& 47.43&  397.1& 1973& 47.39&  398.2\\
    Bulgaria (BG)&              3907& 32.28&  447.9& 2006& 36.04&  444.8&  6699& 52.83&  444.6& 1981& 46.14&  456.9\\
    Czechia (CZ)&               4137& 39.04&  356.9& 2002& 18.78&  386.9&  6744& 47.75&  376.2& 1965& 42.85&  406.5\\
    Denmark (DK)&               3069& 57.09&  457.2& 2015& 56.58&  465.7&  5493& 37.19&  498.8& 1971& 47.39&  529.7\\
    Estonia (EE)&               2595& 36.42&  243.6& 2012& 38.92&  247.5&  {-} &  {-} &   {-}& {-} & {-}  &    {-}\\
    Spain (ES)&                 4770& 44.93&  938.2& 2003& 53.77&  956.3&  7198& 29.27&  935.7& 1930& 40.93&  960.5\\
    Catalonia (ES-CT)&          2077& 46.61&  915.2& 2007& 47.53&  921.0&  1525& 34.82&  896.0& 1999& 35.27&  904.1\\
    Galicia (ES-GA)&             943& 54.08& 1072.1& 2010& 58.16& 1144.2&   953& 42.50& 1138.0& 2000& 43.45& 1164.0\\
    Basque Country (ES-PV)&     {-} &  {-} &    {-}& {-} & {-}  &   {-}&   1031& 43.65&  962.6& 1989& 46.25&  981.9\\
    Finland (FI)&               1179& 42.66&  233.2& 2001& 45.53&  219.8&  6111& 55.41&  227.3& 1986& 49.55&  219.3\\
    France (FR)&                3618& 30.21&  275.3& 2002& 28.17&  292.8&  9813& 62.96&  272.3& 1996& 66.48&  275.3\\
    Great Britain (GB)&        24239& 48.83&  438.5& 2017& 44.67&  465.9& 33257& 43.61&  455.0& 1996& 31.86&  485.7\\
    Greece (GR)&                5639& 46.91&  959.8& 2013& 56.73&  959.7&  6389& 37.30&  971.0& 1972& 42.75&  966.4\\
    Croatia (HR)&               8322& 22.76&  489.7& 2016& 26.88&  504.2& 10741& 60.27&  503.9& 1989& 58.82&  525.8\\
    Hungary (HU)&               2935& 24.16&  581.3& 2020& 24.01&  633.0&  2597& 59.11&  598.8& 2000& 57.70&  585.7\\
    Iceland (IS)&                536& 47.95&  470.0& 2015& 38.26&  552.5&  {-} &  {-} &   {-}& {-} & {-}  &    {-}\\
    Italy (IT)&                 3367& 38.25&  696.5& 2014& 45.78&  707.4&  7848& 62.51&  671.7& 1971& 56.77&  704.5\\
    Latvia (LV)&                 798& 21.30&  357.9& 2008& 19.52&  303.9&  1410& 66.95&  317.5& 1990& 70.50&  303.3\\
    The Netherlands (NL)&       5657& 38.43&  502.5& 2001& 37.83&  473.0&  7906& 58.50&  484.5& 1986& 59.37&  500.7\\
    Norway (NO)&               10998& 50.42&  457.1& 2009& 40.82&  475.7&  {-} &  {-} &   {-}& {-} & {-}  &    {-}\\
    Poland (PL)&                5489& 11.09&  356.4& 2014& 16.88&  359.6&  9705& 45.20&  329.8& 2000& 46.30&  340.1\\
    Portugal (PT)&              3464& 57.74&  459.3& 2001& 56.07&  464.9&  7692& 58.68&  458.6& 1958& 43.16&  451.9\\
    Serbia (RS)&                9914& 16.13&  652.9& 2015& 14.14&  594.5& 15114& 72.91&  650.4& 1990& 65.68&  659.2\\
    Sweden (SE)&                8425& 46.30&  675.2& 2011& 47.39&  702.1&  {-} &  {-} &  {-} & {-} & {-}  &    {-}\\
    Slovenia (SI)&              2726& 73.40&  516.4& 2002& 63.54&  519.5&  9040& 62.52&  533.6& 2014& 49.70&  526.7\\
    Turkey (TR)&               16138& 41.81&  410.3& 2008& 45.72&  413.7& 17384& 48.62&  418.5& 1990& 44.52&  430.3\\
    Ukraine (UA)&               2545& 16.19&  232.3& 2001& 14.79&  242.4& 11324& 68.79&  224.5& 2182& 35.61&  233.3\\
\bottomrule
  \end{tabular}}
  \caption{Statistics of the dataset.
    For each dataset, the number of speeches (n),
    the class imbalance (L\%~--~the percentage of \emph{left} for orientation,
    O\% -- the percentage of \emph{opposition} for power),
    and the average number of tokens are reported.
  }\label{tbl:main-stats}\vspace{-3mm}
\end{table*}%


The data is a subset of ParlaMint version 4.0 \cite{parlamint-v4}.
For the shared task,
we split the data into training and test sets
(without a fixed validation set),
and share them via \url{https://zenodo.org/records/10450640}.
We also provide English translations provided
in the ParlaMint distribution \cite{parlamint-v4en}.
The main motivation for the subsampling is
to reduce the effects of covariates explained above.
Furthermore, since
ParlaMint contains over \num{1.2} billion words, and more than \num{7.7} million speeches
(more correctly `utterances' in ParlaMint TEI annotations),
sampling also results in a more manageable dataset for machine-learning experiments,
promoting inclusion of participants without access to high-performance
computing facilities.

Before sampling the speeches,
we join the utterances by the same speaker
when they were interrupted by a single utterance of another speaker,
and we filter out speeches that are shorter than \num{500} characters,
and longer than \num{20000} characters.
The former is intended for the inclusion of the interrupted speeches 
as a whole.%
\footnote{It is common for the speeches to be interrupted by the chair,
often asking the speaker to finish in the allotted time.
Unauthorized interruptions from the audience are also common.}
The latter, filtering by size,
removes short interruptions and very long speeches.
On average, the lengths of the selected speeches are
between \num{200} and \num{1000} words,
approximately corresponding to speeches of \num{2} to \num{10} minutes.
The utterances of the session chairs,
which are typically about procedural matters,
are always filtered out.

The only preprocessing steps we apply are
replacing the party names or abbreviations as listed in ParlaMint with a placeholder \verb+<PARTY>+,
and using a \verb+<p>+ tag to indicate paragraph boundaries in the original transcripts.
Masking the party references eliminates some trivial cues, as in 
`I am speaking on behalf of \verb+<PARTY>+'.
We only replace the party names and abbreviations
as given in ParlaMint metadata,
which do not cover some of the alternative
names or abbreviations of the parties, 
as well as (consistent) mistranslations
in the automatically translated texts.
We leave the rest of the named entities intact.
Even though (stance towards) some of the named entities may also provide strong 
cues for power and ideology,
many of these cues will be legitimate, and
we expect the models to discover and make use of them
(e.g., the stance towards a particular event, like Brexit,
may genuinely stem from a speakers' relation with the government or their political orientation).
Future releases of the data may improve on eliminating
the obvious cues for power or ideology.

We also include the sex of the speaker,
an anonymised speaker ID,
and automatic translation to English in the training data.
The gender information in ParlaMint was collected from various sources,
typically from the information provided on the web pages of the parliaments,
or from Wikipedia, while 
in a small number of cases, the gender is unknown.
Similarly,
the machine translations are also not available in a small number of instances,
mostly due to technical problems. 
The motivation for including speaker ID is
to provide informed ways of dividing the available data 
as training and validation sets.
The speaker ID is not included in the test set.

\paragraph{Sampling}\label{sec:sampling}
For ideal datasets for both tasks,
we would need a large variation with respect to political party affiliations
and speaker identities.
For example, we would want multiple disjoint left-wing and right-wing
political parties to be present in the training set and the test set
so that the models could be evaluated for their ability to predict
political orientation without relying on party affiliation.
However, the nature of the ParlaMint data
(in fact, any realistic corpus of parliamentary debates)
prevents having such a dataset.
For many parliaments,
the number of political parties of a particular orientation is limited
to a small number.
For the power identification tasks, this is even more severe
since a single party or only a few parties are in power
in some countries
throughout the time period covered in ParlaMint.

As a trade-off between data size, and for reducing the effect of covariates,
we opt for a speaker-based sampling.
First, to discourage, to some extent, the classifiers from relying on author identification,
we sample maximally \num{20} speeches of a single speaker.
This is also important for introducing variation into the dataset,
as the number of speeches from each speaker
follows a power-law distribution.
While a small number of speakers tend to deliver most of the speeches,
e.g., party or party group leaders,
most speakers have relatively few speeches.
The distribution of speeches or speakers to include
in training and test sets is also important for proper evaluation.
For the ideology task, the set of speakers
in the training and test sets are disjoint.
For a reasonably accurate evaluation,
we set the test set size to \num{2000} instances
(about \num{100} to \num{200} speakers depending on the individual corpus and the task).
Despite multiple speeches from each speaker,
due to missing annotations and the lack of diversity of orientation
in some parliaments,
the disjoint training/test constraint above results
in a small number of training instances,
leaving a small number of instances in the training set
for some of the parliaments.

Ideally, power identification requires a different constraint.
That is, the same speaker should be present in both
training and test sets
such that speeches from one set should be when the speaker was in power,
and the other set should contain the speeches while the same speaker is part of the opposition.
This constraint is too difficult, or impossible,
to satisfy for many parliaments in the ParlaMint data.
For example, in Poland,
only a single party is in power throughout the period covered by the corpus.
Similarly,
even when there is some variation,
only a small number of speakers often serve
both in governing coalitions and opposition.
As a result,
we use a best-effort train--test split, where if possible,
we make sure that the speakers in the test set
are also available in the training set with the opposite power role.%
\footnote{The data from only three parliaments (AT, SI, UA) satisfy
this constraint,
while there are no speakers that changed their roles
in ES-GA, HU and PL.}
Otherwise, we randomly sample more speakers to
obtain approximately \num {2000} instances in the test set.
Political systems in some countries do not have a formal
coalition--opposition distinction.
As a result, we leave these parliaments
out of the dataset.

\paragraph{Statistics}\label{sec:stats}
The procedure described above results in training sets from
\num{28} parliaments for the ideology identification task,
and \num{25} parliaments for the power identification task.
Table~\ref{tbl:main-stats} 
provides some statistics on the training and test datasets.
In general, there is a varying class imbalance in both datasets,
but class distribution
and speech lengths between training and test sets are similar.
For some parliaments,
the sampling procedure results in rather small training sets.
Better classification of these datasets
may be achieved by techniques like cross-lingual transfer
and data augmentation.

\begin{table}
  \centering
  \begin{tabular}{l*{4}{@{\hspace{5pt}}S[table-format=3.1,round-precision=1]}}
    \toprule
    ~ & \multicolumn{2}{c}{Orientation} & 
        \multicolumn{2}{c}{Power} \\
    \cmidrule(lr){2-3}\cmidrule(lr){4-5}
    ~      & {dev}   & {test} &
             {dev}   & {test} \\
    \midrule
    AT     & 59.1408 & 51.9033& 68.4700 & 64.9713\\
    BA     & 42.3672 & 41.5554& 45.9694 & 45.8549\\
    BE     & 55.6192 & 56.7218& 58.2705 & 63.3871\\
    BG     & 53.7067 & 53.6677& 61.8393 & 64.7294\\
    CZ     & 54.0443 & 51.1344& 59.0143 & 62.0418\\
    DK     & 50.9093 & 53.9660& 51.6617 & 53.4411\\
    EE     & 47.5340 & 47.4353& {-}     & {-}    \\
    ES     & 72.1190 & 71.7321& 61.2410 & 64.9618\\
    ES-CT  & 72.7830 & 66.4125& 68.6123 & 76.7284\\
    ES-GA  & 62.3931 & 70.5219& 74.3203 & 70.7121\\
    ES-PV  &  {-}    & {-}    & 66.3209 & 68.9317\\
    FI     & 59.3719 & 52.5583& 55.8585 & 52.1178\\
    FR     & 43.9157 & 44.9594& 64.1253 & 66.0806\\
    GB     & 75.8550 & 74.9447& 74.4394 & 70.8660\\
    GR     & 72.5131 & 75.2164& 66.8618 & 64.0252\\
    HR     & 43.7754 & 43.1549& 60.1865 & 59.4355\\
    HU     & 56.2348 & 55.8222& 81.8031 & 84.9224\\
    IS     & 41.5570 & 46.1643& {-}     & {-}    \\
    IT     & 57.2966 & 50.9049& 46.9875 & 43.8796\\
    LV     & 42.8132 & 44.5763& 42.0349 & 52.3242\\
    NL     & 51.4408 & 54.3669& 60.8519 & 64.4872\\
    NO     & 60.9369 & 62.9770& {-}     & {-}    \\
    PL     & 46.3637 & 45.3905& 74.6034 & 75.5809\\
    PT     & 61.7054 & 63.7293& 67.5192 & 63.3542\\
    RS     & 47.8793 & 51.6131& 69.6544 & 62.6876\\
    SE     & 75.5290 & 75.4702& {-}     & {-}    \\
    SI     & 44.5464 & 40.7140& 53.1031 & 53.7169\\
    TR     & 85.7794 & 83.5662& 84.4039 & 81.8516\\
    UA     & 56.7069 & 58.9339& 59.4366 & 45.4016\\
    \bottomrule
  \end{tabular}
  \caption{Macro-averaged F1-scores of the baseline on (dev)elopment and test sets
    on all development and test sets.
    All scores are averages of five random splits of the provided training data
    as \SI{80}{\percent} for training and \SI{20}{\percent} for validation.
    The scores above were obtained without any hyperparameter tuning.
  }\label{tbl:baseline-resutls}
\end{table}
\section{Baselines}

The main purpose of this paper is to introduce the dataset.
However, we also report results from a simple baseline
which is provided for the shared task.
The baseline uses TF-IDF weighted character n-gram features
with a simple logistic regression classifier.
The motivation for such a simple baseline is twofold.
First, since it will be used as the baseline for the shared task,
a competitive baseline may intimidate some of the potential
participants, particularly students and early researchers.
Second, since the baseline only uses `surface' features,
with no claim of `language understanding',
it also provides initial data about how much
of `the politics is about the words'.

Table~\ref{tbl:baseline-resutls} presents the
F1-scores of the baseline for both tasks and for all parliaments.
Most scores are better than a random baseline
(which would result in a \SI{50}{\percent} F1-score). 
Most of the lower scores are the result of relatively high precision and low recall,%
\footnote{Since F1-score favours similar precision and recall values.}
clearly showing the lack of hyperparameter tuning.
The mild correlation between the F1-scores and the training set size
(\num[round-precision=2]{.52791171} and \num[round-precision=2]{.35663038} on orientation and power detection tasks respectively)
and weak but significant correlation of the class imbalance and the scores
(\num[round-precision=2]{-.21017988} and \num[round-precision=2]{-.1613553} on orientation and power detection tasks respectively)
also indicate that the data size and class imbalance
are important factors for the success of the present classifier.
However, these are not the only sources of difficulty.
Despite relatively large datasets,
for example,
AT 
and NO are classified rather poorly for political orientation
(and also the F1-score drops substantially in the test set compared to the development set),
which may be because of better separation of speakers across training and test sets.
On the other hand,
the success of the baseline on both tasks on TR
is unlikely to be explainable by the size and the class imbalance.
One can perhaps relate these to political polarization,
rather than the technical reasons we list above.%
\footnote{A proper investigation of this is beyond
the scope of the current paper.
Hence this statement should only be taken as a potential
future direction for research.}

\section{Conclusions}

The paper presents a dataset derived from the ParlaMint corpora, 
meant for studying automatic methods for detecting political orientation and power position in parliamentary debates.
We believe it could be a valuable resource
for studying these phenomena and
other aspects of political discourse in multiple political and parliamentary
cultures/traditions, and in multiple languages.
Since measuring power and ideology on an individual basis is difficult,
we use the well-known sources of party orientation and power position information
to label individual speeches.
This introduces some strong covariates of the ideology and power
in any dataset that is derived from existing resources.
Instead of a more restrictive setting where covariates are more strictly eliminated,
we opted for a more inclusive dataset of including many parliaments and languages. 
We intend to improve the existing dataset by increasing its coverage and
quality and by adding more metadata.

\section{Limitations}

The orientation and power based on party affiliation
may not always reflect the individuals' positions at the time of 
 their speeches.
However, this is unlikely to be resolved easily
without restricting the number of speakers drastically.
A possible solution, as suggested by an anonymous reviewer,
is to do manual annotations of the individual politicians by the experts,
which would definitely be costly, and may also have its own limitations
, such as changing positions in time.

We did not include the centre even though it clearly falls within
the left--right spectrum of political orientation.
This decision was motivated by simplicity.
The inclusion of a centre in a binary classification scheme is not trivial,
and not all parliamentary corpora include parties annotated as centre.
For the future, multi-class classification,
or a form of ordinal regression/classification may be
interesting alternatives against this limitation.

In the current version of the data, some procedural aspects of speech 
may also provide trivial, unwanted, cues for power and orientation.
More rigorous identification and elimination of these cues
in a big multilingual corpus is a difficult undertaking,
that we leave for a potential new version of the corpus.

\section{Acknowledgements}

This work has been supported by CLARIN ERIC, ParlaMint: Towards Comparable Parliamentary Corpora.

\section{Bibliographical References}\label{reference}
\label{main:ref}

\bibliographystyle{lrec-coling2024-natbib}
\bibliography{parlamint}

\end{document}